\documentclass[letterpaper]{article} 
\usepackage{aaai25}  
\usepackage{times}  
\usepackage{helvet}  
\usepackage{courier}  
\usepackage[hyphens]{url}  
\usepackage{graphicx} 
\usepackage{natbib}  
\usepackage{caption} 
\usepackage{algorithm}
\usepackage{algorithmic}

\usepackage{newfloat}
\usepackage{listings}
\DeclareCaptionStyle{ruled}{labelfont=normalfont,labelsep=colon,strut=off} 
\floatstyle{ruled}
\newfloat{listing}{tb}{lst}{}
\floatname{listing}{Listing}
\usepackage{algorithm}
\usepackage{algorithmic}
\usepackage{amssymb}
\usepackage{amsmath}
\usepackage{array}
\usepackage{booktabs}
\usepackage{tabularx}
\usepackage{multirow}
\usepackage{makecell}

\title{Ultra-High-Definition Dynamic Multi-Exposure Image Fusion \\via Infinite Pixel Learning}
\author{
    Xingchi Chen\textsuperscript{\rm 1,2}, 
    Zhuoran Zheng\textsuperscript{\rm 1,2}\equalcontrib, 
    Xuerui Li\textsuperscript{\rm 3}, 
    Yuying Chen\textsuperscript{\rm 1}, 
    Shu Wang\textsuperscript{\rm 4}, 
    Wenqi Ren\textsuperscript{\rm 1}\equalcontrib
}
\affiliations{

	\textsuperscript{\rm 1}Shenzhen Campus of Sun Yat-sen University, \textsuperscript{\rm 2}Jimei University, \\
	\textsuperscript{\rm 3}The State University of New York at Buffalo, 
	\textsuperscript{\rm 4}Fuzhou University\\
	chenxch66@mail2.sysu.edu.cn, zhengzhr@mail.sysu.edu.cn, \\
	xueruili@buffalo.edu, elvin.chenyuying@gmail.com, \\
	shu@fzu.edu.cn, renwq3@mail.sysu.edu.cn
    
}

\usepackage{bibentry}

\begin{document}

\maketitle

\begin{abstract}
With the continuous improvement of device imaging resolution, the popularity of Ultra-High-Definition (UHD) images is increasing.
Unfortunately, existing methods for fusing multi-exposure images in dynamic scenes are designed for low-resolution images, which makes them inefficient for generating high-quality UHD images on a resource-constrained device. 
To alleviate the limitations of extremely long-sequence inputs, inspired by the Large Language Model (LLM) for processing infinitely long texts,  we propose a novel learning paradigm to achieve UHD multi-exposure dynamic scene image fusion on a single consumer-grade GPU, named Infinite Pixel Learning (IPL). 
The design of our approach comes from three key components: 
The first step is to slice the input sequences to relieve the pressure generated by the model processing the data stream; 
Second, we develop an attention cache technique, which is similar to KV cache for infinite data stream processing; 
Finally, we design a method for attention cache compression to alleviate the storage burden of the cache on the device. 
In addition, we provide a new UHD benchmark to evaluate the effectiveness of our method. 
Extensive experimental results show that our method maintains high-quality visual performance while fusing UHD dynamic multi-exposure images in real-time ($>$40fps) on a single consumer-grade GPU.
\end{abstract}

%
\begin{figure}[t]
	\centering
	\includegraphics[width=\columnwidth]{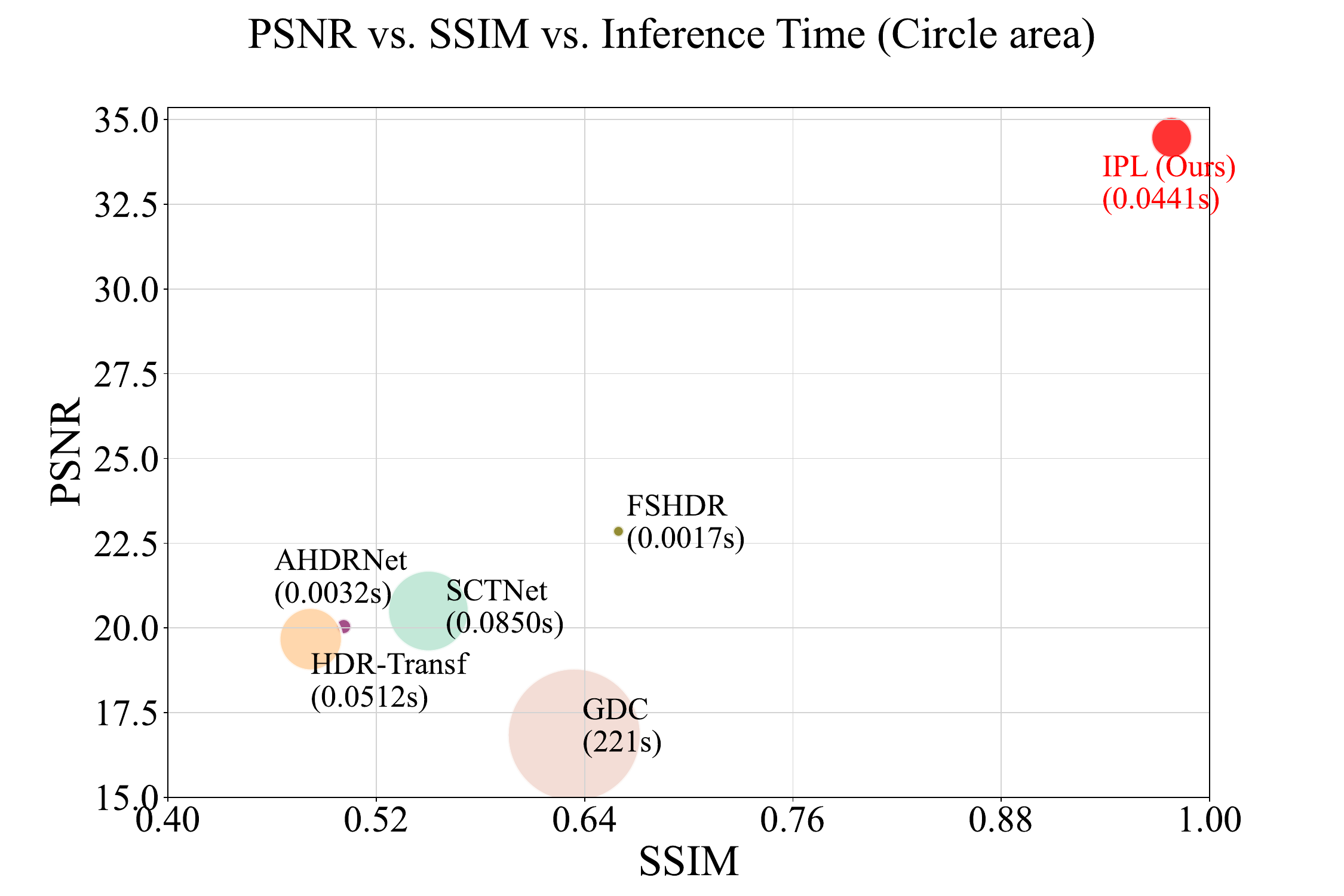} 
	\caption{Model performance and efficiency comparison between our proposed IPL and other MEF methods on our proposed dataset. Since most methods are unable to process UHD images directly, the calculation is performed based on the maximum resolution~\cite{U:16} that these algorithms can handle on a single GPU. Our method has approximately \textbf{46\%} higher PSNR and \textbf{48\%} higher SSIM than the second-best method, and the inference speed reaches real-time ($>$40fps), achieving an optimal trade-off between performance and efficiency.}
	\label{fig1}
\end{figure}
\section{Introduction}

Using multiple images with different exposures to generate a clear image is a common technique.
In recent years, due to the emergence of sophisticated imaging sensors and displays, the popularity of Ultra-High-Definition (UHD) images is rapidly increasing. 
However, existing multi-exposure image fusion (MEF) algorithms focus on generating a single low-resolution image and cannot run multiple UHD images with different exposures on resource-constrained devices.

To date, many methods have been developed to enable running UHD images on consumer GPUs, such as UHD image dehazing~\cite{U:16}, UHD deblurring~\cite{U:17}, and UHD deraining~\cite{UHDderaining}.
However, most of these methods focus on processing a single image with 4K ($3840 \times 2160$) resolution or higher on a single consumer GPU, which is usually inefficient for processing multiple heterogeneous UHD images.
In this research field, among existing multi-exposure fusion algorithms, such as traditional algorithms~\cite{c:1,c:2,c:4}, only a few of them can process UHD images in real time.
Unfortunately, the images obtained by manually extracting features are prone to artifacts or ghosting.
In addition, among deep learning-based methods~\cite{c:8,c:9,c:10,c:11,c:12,BracketIRE} only AHDRNet~\cite{c:8}, FSHDR~\cite{c:10}, and BracketIRE~\cite{BracketIRE} can process UHD images in real-time, but they generate UHD images with low quality. 
In this paper, we attempt to develop an efficient deep network that balances speed and accuracy (see Figure~\ref{fig1}).

Inspired by Large Language Models (LLMs) for processing extremely long sequential inputs \cite{L:1}, we highlight the importance of chunking and cache acceleration. 
%
To boost the inference speed of the model, LLMs leverage KV caches, which store precomputed key/value vectors and reuse them for token generation, reducing redundant computations~\cite{L:2}. 
Building on this foundation, ChunkAttention~\cite{L:3} enhances self-attention speed using fragmentation, shared KV caches, and batched attention to reduce memory operations. FastGen~\cite{L:2} enables adaptive KV cache construction through lightweight attention analysis. 
%
%
Obviously, in UHD image processing, adopting chunk-cache techniques is crucial for accelerating speed while ensuring accuracy. 
%
Additionally, it is worth noting that KV cache memory consumption increases rapidly with model size and generation length, placing a significant burden on device memory. 
Multiple studies \cite{L:4,L:5,L:6} aim to optimize memory usage and computational efficiency of LLMs, which is crucial for resource-constrained environments and real-time applications. 
Therefore, we develop a new machine learning paradigm based on the chunk-cache-quantization pipeline, i.e. infinite pixel learning.

Specifically, in the Dimensional Attention Enhancement Module (DAEM), 
we first chunk the input data along the channel, width, and height dimensions, and then apply the cyclic scanner to extract and enhance important details in each dimension.  
Second, we develop an attention cache technique, which is similar to a KV cache for infinite data stream processing. The attention cache stores the extracted features to prevent redundant calculations and accelerate inference. 
Finally, we employ quantization compression to reduce the storage burden of the cache on the device. 
Next, we propose the Dimensional Rolling Transformation Module (DRTM) to address the loss of overall features during the cyclic scanner process. 
Inspired by, yet distinct from, MLP-Mixer~\cite{MPL}, DRTM encodes the feature maps of the image from the perspective of channel, width, and height.  
By arranging feature maps from different views, DRTM effectively models long-range dependencies. 
Furthermore, we provide a UHD benchmark to evaluate the effectiveness of our method. 

Our main contributions are summarized as follows: 
\begin{itemize}
	\item We develop a novel learning paradigm named \textbf{Infinite Pixel Learning (IPL)}, which processes infinite continuous data streams using three key components: slice cyclic scanner, attention cache technique, and quantization compression. 
	IPL enables full-resolution UHD image inference on a single consumer-grade GPU while effectively avoiding ghosting artifacts. 
	
	\item We design an efficient network that contains two main modules: one for extracting the local features of the image and the other for capturing the global features of the image through dimensional scrolling transformation.
	
	\item We introduce 4K-DMEF, the first UHD MEF benchmark dataset for dynamic scenes. Our method, tested on both this UHD dataset and other non-UHD datasets, achieves a balance between performance and efficiency.
\end{itemize}

\begin{figure*}[t]
	\centering
	\includegraphics[width=2.1\columnwidth]{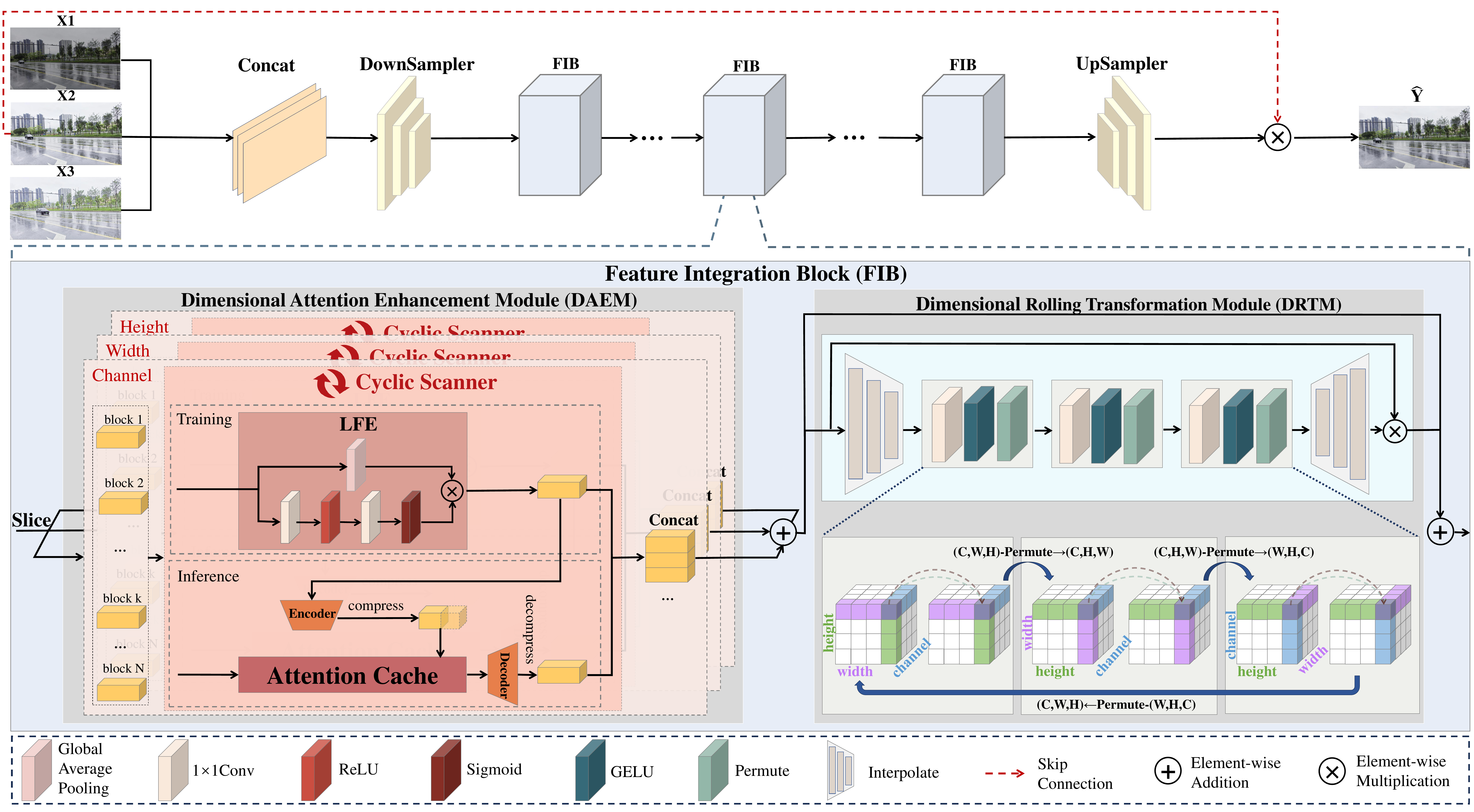} 
	\caption{The overall architecture of IPL, which extracts features using a series of Feature Integration Blocks (FIBs).   
		The FIB mainly contains a Dimensional Attention Enhancement Module (DAEM) and a Dimensional Rolling Transformation Module (DRTM). 
		DAEM has three key components: Slice Cyclic Scanner, Attention Cache Technique, and Quantization Compression, forming a chunk-cache-quantization pipeline to process infinite input pixels efficiently.
		DRTM associates features from different views by permuting feature maps to compensate for global features.}
	
	\label{fig2}
\end{figure*}
\section{Related Works}
\noindent \textbf{Dynamic Multi-Exposure Image Fusion.}
Multi-exposure image fusion technology tackles detail loss in highlights and shadows of single-exposure images, preserving detail across all exposures.  
However, ghosting artifacts from moving objects or camera misalignment remains a key challenge in dynamic scenes  \cite{tip21,tip23}. 
To avoid ghosting artifacts, AHDRNet~\cite{c:8} uses the attention module to guide merging according to the reference image. 
FSHDR~\cite{c:10} proposed pioneering work to achieve efficient ghosting removal using zero-shot and few-shot learning strategies. 
HDR-Transformer~\cite{c:11} and SCTNet~\cite{c:12} utilize transformers to capture key factors for ghosting removal. 
Although several of the above studies have made significant progress, they have primarily focused on low-resolution images. 
When applied to UHD images, these methods often encounter memory overflow issues and are unable to perform full-resolution inference on a single consumer-grade GPU.

\noindent \textbf{Ultra-High-Definition Image Processing.}
In recent years, advancements in sensors and displays have rapidly increased the popularity of UHD images. 
Consequently, many advanced methods have been developed to handle UHD images on consumer GPUs.  
Existing UHD image processing methods are primarily employed in various applications such as dehazing~\cite{U:16}, deblurring~\cite{U:17}, deraining~\cite{UHDderaining,De-Raining_Transformer} and low-light enhancement \cite{U:18,U:19}. 
These methods are designed to improve image quality in specific challenging conditions. 
Additionally, UHDFormer ~\cite{U:20} is the first universal UHD image restoration Transformer,  advancing the development of UHD image processing technology. 
However, most methods address processing a single 4K ($3840 \times 2160$) or higher resolution image on a consumer GPU, which is often inefficient for handling multiple heterogeneous UHD images.  
Therefore, we focus on fusing UHD dynamic multi-exposure images on a single consumer GPU, marking the first approach to address this task specifically for UHD images. 

\noindent \textbf{Extremely Long Input Sequences Processing.}
As LLMs are increasingly applied to tasks like long document generation and extended dialogue systems, they face challenges with length generalization failures on lengthy text sequences~\cite{L:1}. 
To address these challenges, several techniques have been proposed~\cite{L:2}, including block segmentation, KV cache technology, and memory-efficient architectures. 
ChunkAttention~\cite{L:3} improves self-attention speed by employing fragmentation, a shared KV cache, and batched attention to reduce memory operations.   
\mbox{FastGen}~\cite{L:2} optimizes adaptive KV cache construction through lightweight attention analysis. 
Additionally, multiple studies \cite{L:4,L:5,L:6} focus on optimizing memory usage and computational efficiency in LLMs, which is crucial for resource-constrained environments and real-time applications. 
Inspired by these approaches, we develop a new machine learning paradigm based on a chunk-cache-quantization pipeline to process infinite pixel inputs.

\section{Methodology}
\subsection{Overall Architecture}
An overview of IPL is shown in Figure 2. 
Given three images with different exposures $\mathbf{X_i} \in \mathbb{R}^\mathrm{{3 \times W \times H}},\  \mathbf{i} \mathrm{\in \{1,2,3\}}$, we first concatenate them and map them to the feature space through a DownSampler to obtain the low-level features $\mathbf{F_0}\in \mathbb{R}^\mathrm{{C \times W \times H}}$, where $\mathrm{C}$, $\mathrm{W}$, and $\mathrm{H}$ represent channel, width, and height, respectively. 
Then, the multiple stacked Feature Integration Blocks (FIBs) are used to generate finer deep features $\mathbf{F_i}$ from $\mathbf{F_0}$ for image fusion, where a Feature Integration Block consists of a Dimensional Attention Enhancement Module (DAEM), and a Dimensional Rolling Transformation Module (DRTM). 
Finally, the sum of the final features $\mathbf{F_f}$ is fed to an UpSampler and then multiplied with $\mathbf{X_2}$ to obtain the fused image $\hat{\mathbf{Y}}$.
DownSampler and UpSampler are both composed of a sub-pixel convolutional layer \cite{He_2016_CVPR} and a 3$ \times$ 3 convolutional layer.

\subsection{Dimensional Attention Enhancement Module}
DAEM aims to capture the local features of images. Inspired by LLMs for processing extremely long sequential inputs, we develop a new machine learning paradigm based on the chunk-cache-quantization pipeline, i.e. infinite pixel learning. 
Specifically, DAEM includes a Slice Cyclic Scanner, Attention Cache Technique, and Quantization Compression.

\subsubsection{Slice Cyclic Scanner.}
To alleviate the load on the model when processing the data stream, we follow \mbox{ChunkAttention}~\cite{L:3} and chunk the input data along the three dimensions: channel, width, and height. 
Next, to capture the local features, we apply the cyclic scanner to the blocks in each dimension. The outputs of the three dimensions are added to produce the final output of the DAEM.   
Within the cyclic scanner, we build the local feature extractor (LFE). Specifically, in LFE, 
we simultaneously feed the input blocks into both global average pooling and a series of convolutions to extract deep features. 
These deep features are then processed through a sigmoid function to generate attention weights for the input features. 
Subsequently, the input features are then adjusted according to these weights via element-wise multiplication, enabling dynamic channel importance adjustment. 
The circulation continues until all blocks have been feature extracted by LFE.  
LFE can be described by the following equations:
\begin{align}
	&\mathbf{F_m} = \mathrm{GAP(\mathbf{F_{in}})}, \\
	&\mathbf{F_w} = \mathrm{Sigmoid(Conv(RELU(Conv(\mathbf{F_{in}}))))}, \\
	&\mathbf{F_{LFE}} = \mathbf{F_m} \odot \mathbf{F_w},
\end{align}
where $\mathbf{F_{in}}$ is the input features, $\mathbf{F_m}$ is the averaged features, $\mathbf{F_w}$ is the channel weights, and $\mathbf{F_{LFE}}$ indicates the final output features of LFE.
$\mathrm{GAP(\cdot)}$ denotes global average pooling, $\mathrm{Sigmoid(\cdot)}$ and $\mathrm{ReLU(\cdot)}$ are the Sigmoid and ReLU functions, respectively, $\mathrm{Conv(\cdot)}$ denotes a $1 \times 1$ convolution, and $\odot$ represents the element-wise product.

\subsubsection{Attention Cache Technique.}
To boost the inference speed of the model, LLMs usually employ the KV cache mechanism to trade storage space for reduced inference time~\cite{L:2}. 
Inspired by this approach, we developed a similar technique called attention cache. 
In LFE, deep features are processed through a sigmoid function to generate attention weights for the input features.  
Consequently, the attention cache is used to cache local features extracted by LFE, which helps to avoid redundant calculations and accelerates inference.  
The technique can be expressed as: 
\begin{align}
	\mathbf{F_k}=
	\begin{cases}
		\mathrm{CR(\mathbf{k})}; &\text{if}\ \mathrm{LFE(\mathbf{k})}\ \text{in} \ \mathrm{AC}, \\
		\mathrm{LFE(\mathbf{k})};\  \mathrm{CW(\mathbf{k})}; &\text{otherwise},
	\end{cases}
\end{align}
where $\mathbf{F_k}$ denotes the final local features of the block $\mathbf{k}$, $\mathrm{AC}$ is the attention cache, and $\mathrm{LFE(\cdot)}$ represents LFE. 
$\mathrm{CR(\cdot)}$ and $\mathrm{CW(\cdot)}$ denote the read and write operations in the attention cache, respectively. The write operation will write both the $\mathbf{k}$ value and corresponding $\mathrm{LFE(\mathbf{k})}$ into the attention cache. 

During training, the normal computational intermediate results (feature maps) of the network are computed through convolutional operations. During inference, the convolutional operation is stored as a cache, which is read directly without a computational process, so the whole network can achieve accelerated inference.

\subsubsection{Quantization Compression.}
With model training and inference proceeding, the memory consumption of the attention cache increases rapidly, significantly intensifying the burden of device memory. 
To alleviate this problem, we use quantization compression to reduce storage memory. 

Specifically, we build an encoder to compress the float tensor into the quantized tensor, for a float tensor $\mathbf{t}$ the \mbox{process} can be written as: 
\begin{align}
	\mathbf{s} &= \mathrm{\frac{\mathbf{t_{max}-t_{min}}}{\mathbf{q_{max}-q_{min}}}}, \\
	\mathbf{zp} &= \mathrm{MAX(\mathbf{q_{min}},MIN(\mathbf{q_{max}},Round(\mathbf{q_{min}}-\frac{\mathbf{t_{min}}}{\mathbf{s}})))},\\
	\mathbf{t_q} &= \mathrm{MAX(\mathbf{q_{min}},MIN(\mathbf{q_{max}},Round(\mathbf{\frac{t}{s+zp}})))},
\end{align}
where $\mathbf{s}$ is the scaling factor, $\mathbf{zp}$ is the zero point, and $\mathbf{t_q}$ is the quantized tensor. 
$\mathbf{t_{max}}$ and $\mathbf{t_{min}}$ are the maximum and minimum values in the tensor $\mathbf{t}$, respectively. 
$\mathbf{q_{max}}$ and $\mathbf{q_{min}}$ denote the maximum and minimum values of the quantization range, respectively.  
$\mathrm{MAX(\cdot)}$ and $\mathrm{MIN(\cdot)}$ take the maximum and minimum values of the comparison, respectively. $\mathrm{Round(\cdot)}$ converts a floating-point number to the nearest integer.

Correspondingly, we build a decoder to decompress the quantized tensor, the process can be written as: 
\begin{align}
	\mathbf{t_{d}} &= \mathrm{\mathbf{s} \times (\mathbf{t_q}-\mathbf{zp})},
\end{align}
where $\mathbf{t_{d}}$ denote dequantized tensor.
%
%
%

Furthermore, the process of extracting local features can be written as: 
\begin{align}
	\mathbf{F_k}=
	\begin{cases}
		\mathrm{Decoder(CR(\mathbf{k}))}; &\text{if}\ \mathrm{LFE(\mathbf{k})}\ \text{in} \ \mathrm{AC}, \\
		\mathrm{LFE(\mathbf{k})};\  \mathrm{Encoder(CW(\mathbf{k}))}; &\text{otherwise},
	\end{cases}
\end{align}
where $\mathrm{Encoder(\cdot)}$ denotes quantization compression operation, and $\mathrm{Decoder(\cdot)}$ represents decompression operation.

\subsection{Dimensional Rolling Transformation Module}
To address the loss of overall features during cyclic scanners, we propose DRTM. 
Inspired by, yet distinct from, MLP-Mixer~\cite{MPL}, DRTM encodes the feature maps of the image from the perspective of channel, width, and height. 
To capture long-range dependencies of features, we ingeniously establish relationships across both spatial and channel dimensions using simple dimension transformation operations. 
This approach associates and fuses the information encoded in each feature map through a dimensional rolling transformation operation. 

The architecture of the DRTM and details of dimension transformation operations are shown in Figure~\ref{fig2}. 
Specifically, we first adjust the resolution of the input features and then perform some dimension transformation operations on them. 
Given the input feature $\mathbf{F_{in}}$, this procedure can be written as:
\begin{align}
	&\mathbf{F'_t} = \mathrm{IP(\mathbf{F_{in}})}, \\
	&\mathbf{F_t} =\underbrace{\mathrm{Permute(GELU(Conv(\mathbf{F'_t})))}}_{\times 3}, 
\end{align}
where $\mathbf{F'_t}$ and $\mathbf{F_t}$ are intermediate results, $\mathrm{IP(\cdot)}$ corresponds to the interpolation operation, $\mathrm{Conv(\cdot)}$ is a $1 \times 1$ convolution, $\mathrm{GELU(\cdot)}$ represents GELU function, $\mathrm{Permute(\cdot)}$ denotes the dimension transformation operation and $\times 3$ denotes three operations in sequence. 

Afterward, we use interpolation to adjust the feature $\mathbf{F_t}$ to its original resolution to estimate the attention map 
and adaptively modulate the input $\mathbf{F_{in}}$ according to the estimated attention via element-wise product. This process can be written as:
\begin{align}
	&\mathbf{F_G} = \mathrm{IP(\mathbf{F_t}) \odot \mathbf{F_{in}}},
\end{align}
where $\mathbf{F_G}$ are the final output features and $\odot$ represents element-wise product.

\subsection{Feature Integration Block}
In general, our FIB pipeline can be written as:
\begin{align}
	&\mathbf{F_{DAEM}} = \mathrm{DAEM(\mathbf{\hat{F}_{in}})}, \\
	&\mathbf{F_{out}} = \mathbf{F_{DAEM}} + \mathrm{DRTM(\mathbf{F_{DAEM}})},
\end{align}
where $\mathbf{\hat{F}_{in}}$ is the intermediate features, $\mathbf{F_{DAEM}}$ represents output of DAEM, and $\mathbf{F_{out}}$ denotes the final output features. 

\subsection{Loss Function}
To optimize the weights and biases of the network, we utilize the L1 loss in the RGB color space as the fundamental reconstruction loss. This approach ensures that the network accurately captures and reproduces the fine details and color information of the input images. L1 loss can be written as:
\begin{align}
	&\mathrm{L = \lVert \mathbf{Y} - \mathbf{\hat{Y}} \rVert_1},
\end{align}
where $\mathbf{Y}$ and $\mathbf{\hat{Y}}$ denote the GT and output, respectively.

\begin{figure*}[t]
	\centering
	\includegraphics[width=2.12\columnwidth]{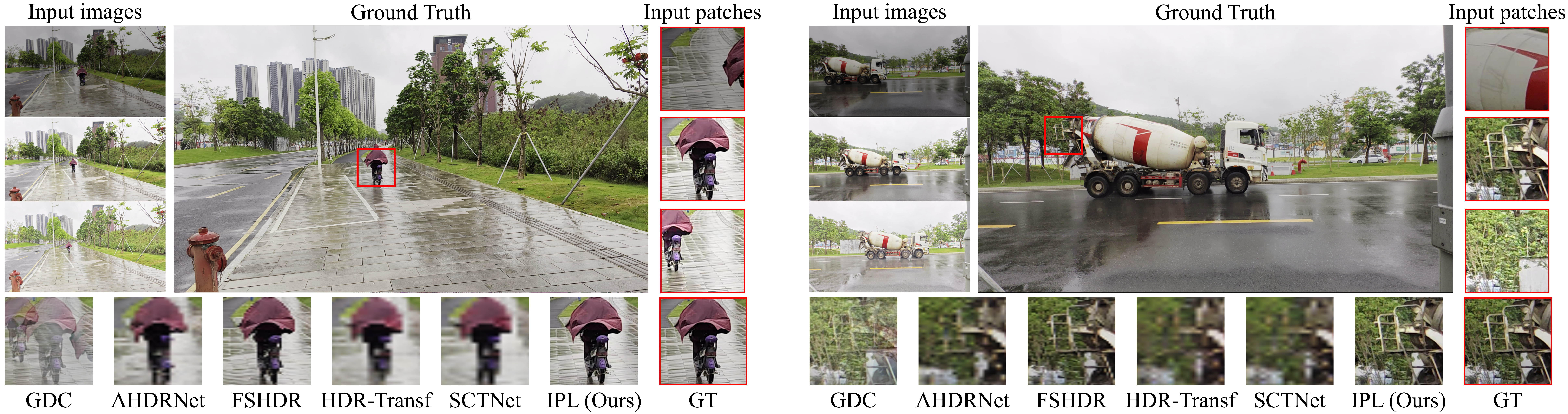} 
	\caption{Qualitative comparison on our proposed 4K-DMEF dataset. 
		All methods are trained using our training set on a single RTX 4090 GPU. 
		Our method, IPL, outperforms all SOTA methods in processing UHD multi-exposure image inputs. 
		It effectively avoids ghosting and blurring, achieving full-resolution inference with remarkable clarity and precision.
	}
	\label{high}
\end{figure*}
\begin{table*}[t]
	\centering
	
	\begin{tabularx}{\textwidth}{l| c| c c c c c c}
		\Xhline{1.2pt}
		\textbf{Methods} & \textbf{MR} & \textbf{PSNR$\boldsymbol{\uparrow}$} & \textbf{SSIM$\boldsymbol{\uparrow}$} & \textbf{LPIPS$\boldsymbol{\downarrow}$} & \textbf{TIME$\boldsymbol{\downarrow}$} &\textbf{MACs}$\boldsymbol{\downarrow}$ &\textbf{Params}$\boldsymbol{\downarrow}$\\
		\Xhline{1pt}
		GDC~\cite{c:1} & 3840 × 2160 & 16.84 & 0.6340 &0.3584  &221.00s &242.20M &- \\
		AHDRNet~\cite{c:8} & 256 × 256& 20.04 & 0.5012 & 0.4906 & 0.0032s&162.77M &1.2398M \\
		FSHDR~\cite{c:10} & 512 × 512& 22.85 & 0.6595 & 0.3520 & \underline{0.0017s}&957.92M&1.8284M\\
		HDR-Transf~\cite{c:11} & 200 × 200& 19.67 & 0.4822 & 0.5202 & 0.0512s&95.94M&1.1978M\\
		SCTNet~\cite{c:12} & 200 × 200& 20.01 & 0.5302 & 0.5204 & 0.0850s&\underline{78.67M}&\underline{0.9615M}\\
		BracketIRE~\cite{BracketIRE} & 1024 × 1024& \underline{27.44} & \underline{0.8651} & \underline{0.2045} & \textbf{0.0007s}&2.61T&\textbf{0.2368M}\\
		IPL (Ours) & 256 × 256& \textbf{33.29}  & \textbf{0.9776} & \textbf{0.0427} & 0.0441s&\textbf{66.95M}&0.9747M\\
		\Xhline{1.2pt}
	\end{tabularx}
	\caption{Comparison of quantitative results on our 4K-DMEF datasets. 
		MR denotes the maximum resolution each algorithm can handle on a single RTX 4090 GPU.  
		TIME represents the inference time required for each algorithm to fuse a single image. 
		The best and second-best values are highlighted in bold and underlined, respectively.}
	\label{tab1}
\end{table*}
\section{Experiments}
In this section, we evaluate the proposed method by conducting comprehensive experiments on both our UHD dataset and several public non-UHD datasets. 
We compare our method against five state-of-the-art multi-exposure fusion (MEF) methods of GDC~\cite{c:1}, AHDRNet~\cite{c:8}, FSHDR~\cite{c:10}, HDR-Transformer~\cite{c:11}, and \mbox{SCTNet}~\cite{c:12}. 
In addition, we conduct ablation studies to show the effectiveness of each module within our network. 
More experimental results are provided in the \mbox{Supplementary Materials}.

\subsection{Datasets}
\subsubsection{Our UHD Dynamic Multi-Exposure Image Dataset.}
To train and evaluate the proposed network as well as the comparison methods, we build a benchmark dataset named 4K-DMEF. 
Specifically, we record videos at a resolution of 4K (3840 $\times$ 2160) with a mobile phone, capturing dynamic elements such as moving people and cars. 
From these videos, we extract three frames from a video, which we label as \mbox{samples} 1, 2, and 3 of dynamic scenes. 
To synthesize different exposure levels, we employ the LECARM~\cite{LECARM} method.
Using LECARM, we composite each of the three frames into three different exposure levels: low, medium, and high. 
For consistency and accuracy in our dataset, we use the original image of sample 2 as the ground truth reference. 
Ultimately, we collected data for 110 UHD dynamic scenes, dividing them into 80 scenes for training and 30 for testing.

\subsubsection{Non-UHD Dataset.}
We also conduct experiments using two public non-UHD dynamic multi-exposure image datasets: Kalantari Dataset~\cite{Kalantari2017DeepHD} and Mobile-HDR Dataset~\cite{Mobile-HDRcvpr23}. 
\mbox{Kalantari} Dataset~\cite{Kalantari2017DeepHD} is a conventional benchmark widely used in previous works~\cite{c:8,c:10,c:11,c:12}.
It has a resolution of 1500 $\times$ 1000 pixels and includes 74 training scenes and 15 test scenes. 
Mobile-HDR Dataset~\cite{Mobile-HDRcvpr23} is a recent dataset captured using mobile phone cameras. 
We divide this dataset into 85 \mbox{training} scenes and 30 test scenes, each with a resolution of 2000 $\times$ 1500 pixels.

\subsection{Implementation Details}
We conduct our experiments using PyTorch on a single NVIDIA GeForce RTX 4090 GPU.  
To optimize the network, we employ the AdamW optimizer with a learning rate $2 \times 10^{-4}$. The network undergoes training for 1200 epochs with a batch size of 4.  
Additionally, the number of Feature Integration Blocks (FIBs) is 8, and the number of feature channels is 48. 
It is important to note that many existing methods are unable to perform full-resolution inference directly on UHD images. 
Inspired by Zheng~\cite{U:16}, for these methods, we first downsample the input UHD images using bilinear interpolation to the maximum resolution that these algorithms can handle on a single GPU. 
After performing inference, we upsample the output back to 4K resolution. 

\begin{figure*}[t]
	\centering
	\includegraphics[width=2.12\columnwidth]{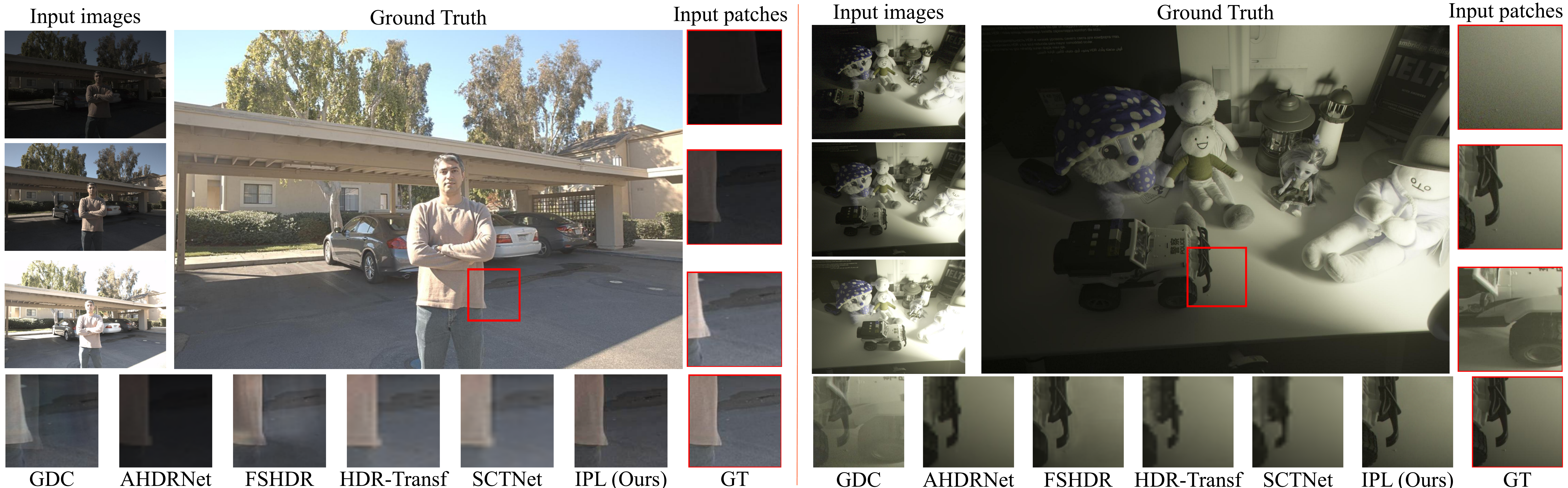} 
	\caption{Visual comparison on the public non-UHD dataset. 
		In the first comparison (left) with the Kalantari dataset~\cite{Kalantari2017DeepHD}, our IPL method shows competitive performance. 
		In the second comparison (right) with the Mobile-HDR dataset~\cite{Mobile-HDRcvpr23}, our IPL method excels in detail restoration and performs well in relatively low-light conditions.
	}
	\label{low}
\end{figure*}

\begin{table}[t]
	\centering
	\small
	\begin{tabularx}{0.46\textwidth}{l| c c}
		\Xhline{1.0pt}
		\textbf{Methods} &\textbf{PSNR}&\textbf{SSIM}\\ 
		\Xhline{1.0pt}
		GDC \cite{c:1} 
		&27.87 & 0.7286 \\
		AHDRNet \cite{c:8} 
		&42.29 &0.9882   \\
		FSHDR \cite{c:10} 
		&41.79& 0.9900 \\
		HDR-Transf \cite{c:11} 
		&42.18& 0.9884 \\
		SCTNet \cite{c:12} 
		&42.49& 0.9887 \\
		IPL (Ours) 
		&42.13 &0.9902   \\
		\Xhline{1.0pt}
	\end{tabularx}
	\caption{Comparison of quantitative results on Kalantari.}
	\label{tab2}
\end{table}

\subsection{Comparisons with State-of-the-Art}
\subsubsection{Quantitative Results\\}
We employ three well-known image quality assessment metrics, namely PSNR, SSIM~\cite{P-S-metrics}, and LPIPS~\cite{lpips}, to quantify the performance of different methods. 
Also, we record the highest resolution that each algorithm can handle on a single RTX 4090 GPU, referred to as Maximum Resolution, and the time it takes to generate a fused image, referred to as Inference Time.

In Table~\ref{tab1}, we present the quantitative comparison results of various methods evaluated on our proposed 4K-DMEF dataset. 
It can be observed that our IPL method achieves approximately 46\% and 48\% higher PSNR and SSIM, respectively, compared to the second-best method, FSHDR. 
Notably, a lower LPIPS value indicates better perceptual quality. Our IPL method also shows a significant improvement in LPIPS, achieving a value approximately 88\% better than that of FSHDR. 
For inference time, although our method is not the fastest among the compared approaches, it still achieves real-time inference at more than 40 fps.
Tables~\ref{tab2} and~\ref{tab3} show that our IPL achieves performance comparable to the state-of-the-art methods on non-UHD datasets. 
Combined with the results shown in Figure~\ref{low}, these findings indicate that our IPL method maintains high-quality results and performs competitively across different resolutions of multi-exposure image fusion.
\begin{table}[t]
	\centering
	\small
	\begin{tabularx}{0.46\textwidth}{l| c c}
		\Xhline{1.2pt}
		\textbf{Methods} &\textbf{PSNR} &\textbf{SSIM}\\
		\Xhline{1pt}
		GDC \cite{c:1} 
		&28.02 &0.6148\\
		AHDRNet \cite{c:8} 
		&36.64& 0.9803 \\
		FSHDR \cite{c:10} 
		&32.56&0.9706  \\
		HDR-Transf \cite{c:11} 
		&36.67&0.9809  \\
		SCTNet \cite{c:12} 
		&36.20&0.9788  \\
		IPL (Ours) 
		&36.45  &0.9902   \\
		\Xhline{1.2pt}
	\end{tabularx}
	\caption{Comparison of quantitative results on Mobile-HDR.}
	\label{tab3}
\end{table}

\subsubsection{Qualitative Results\\}
In addition to the quantitative evaluations, we provide qualitative comparisons to further illustrate the effectiveness of the proposed method. 
Figure~\ref{high} presents the visual comparisons, providing a detailed view of the results produced by our method compared to other MEF approaches. 

As shown in Figure~\ref{high}, the fusion results generated by the traditional GDC~\cite{c:1} method frequently exhibit ghosting artifacts.  
Additionally, the fusion results from other neural network-based methods tend to be blurred to varying degrees. 
This blurring occurs because these methods are unable to directly process UHD images with different exposures on a single GPU, leading to a loss of pixel information during the necessary downsampling process. 
Obviously, our IPL approach is the only method that can efficiently perform full 4K resolution inference on a single GPU, avoiding common issues such as ghosting and blurring. 
It effectively fuses UHD dynamic multi-exposure images, overcoming the challenges that other methods \mbox{typically face}.

\subsection{Ablation study}
We perform three ablation studies to demonstrate the effectiveness of each IPL component, evaluating each in a fair setting. 
For these experiments, we use the same architecture and hyperparameters, varying only one component in each ablation. 
The evaluation of these ablation experiments is conducted on our 4K-DMEF dataset.

\begin{table}[t]
	\centering
	\small
	\scalebox{0.92}{
		\begin{tabularx}{0.5\textwidth}{c|c c c|c|c c c}
			\Xhline{1pt}
			\multirow{2}{*}{\textbf{Group}} & \multicolumn{3}{c|}{\textbf{DAEM}} & \multirow{2}{*}{\textbf{DRTM}} & \multirow{2}{*}{\textbf{PSNR}$\boldsymbol{\uparrow}$} & \multirow{2}{*}{\textbf{SSIM}$\boldsymbol{\uparrow}$} &\multirow{2}{*}{\textbf{LPIPS}$\boldsymbol{\downarrow}$} \\
			\cline{2-4}
			& \multirow{-0.9}{*}{C} & \multirow{-0.9}{*}{W} & \multirow{-0.9}{*}{H} &  &  &  & \\
			\Xhline{1pt}
			(a)& & & &\checkmark &8.92 &0.5553 &0.4190\\
			(b)&\checkmark & & &\checkmark &32.96 &0.9776 &0.0408\\
			(c)& &\checkmark & &\checkmark &34.44 &0.9776 &0.0403\\
			(d)& & &\checkmark &\checkmark &34.61 &0.9784 &0.0404\\
			(e)&\checkmark &\checkmark & &\checkmark &33.15 &0.9778 & 0.0404\\
			(f)&\checkmark & &\checkmark &\checkmark &32.99 &0.9778 & 0.0412\\
			(g)& &\checkmark &\checkmark &\checkmark &34.51 &0.9782 & 0.0401\\
			(h)&\checkmark &\checkmark &\checkmark & &32.82 &0.9775 & 0.0489\\
			(i)&\checkmark &\checkmark &\checkmark &\checkmark &\textbf{34.78} &\textbf{0.9805} &\textbf{0.0397} \\
			\Xhline{1pt}
		\end{tabularx}
	}
	\caption{Ablation study on the key components of DEAM and DRTM. The comparison is conducted on our 4K-DMEF.}
	\label{tab4}
\end{table}

\begin{figure}[t]
	\centering
	\includegraphics[width=\columnwidth]{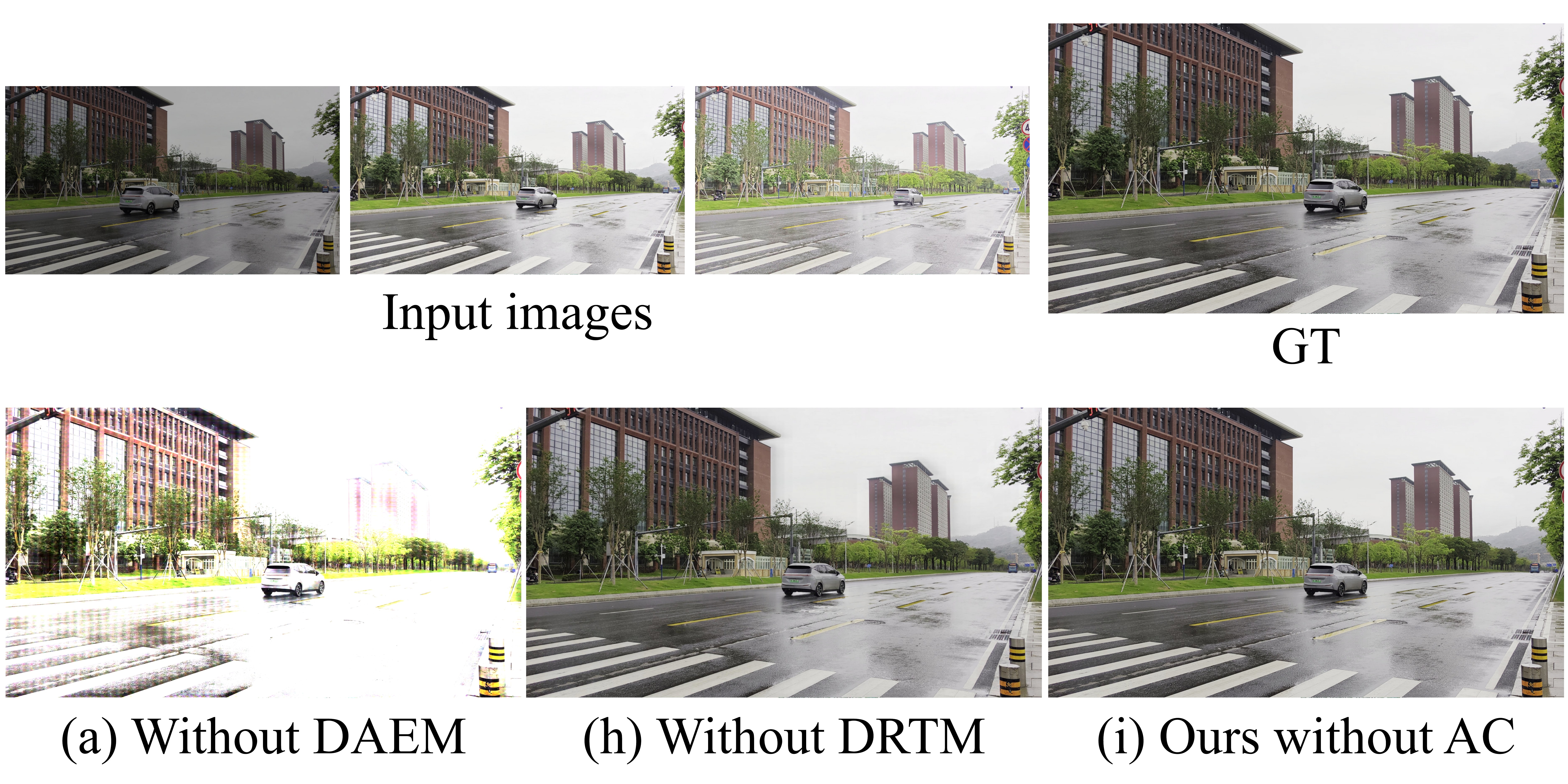} 
	\caption{Visualization results of ablation Experiments. 
		As shown, without DAEM, the model suffers severe performance degradation, failing to capture overall features and resulting in a slightly different color tone from the GT.
	}
	\label{vab}
\end{figure}

\begin{table}[t]
	\centering
	
	\begin{tabularx}{0.41\textwidth}{c| c c c}
		\Xhline{1pt}
		\textbf{Attention Cache} & \textbf{PSNR}$\boldsymbol{\uparrow}$ & \textbf{SSIM}$\boldsymbol{\uparrow}$ & \textbf{TIME}$\boldsymbol{\downarrow}$ \\
		\Xhline{1pt}
		& 34.78 & 0.9805& 0.1475s \\
		\checkmark & 33.29& 0.9776 & \textbf{0.0441s} \\
		\Xhline{1pt}
	\end{tabularx}
	\caption{Ablation study on the key components of Attention Cache. The comparison is conducted on our 4K-DMEF dataset. TIME denotes the time taken to generate an image.}
	\label{table5}
\end{table}
\noindent \textbf{Effectiveness of Dimensional Attention Enhancement Module.}
Based on the results in Table~\ref{tab4}, it is obvious that, without the DAEM, the model experiences the most severe performance degradation (see Figure~\ref{vab}). 
This aligns with our hypothesis, as the primary mechanisms for handling extremely long pixel input are crucial factors contributing to the outstanding performance of IPL. 
From the results of group (b) to (h), it is evident that all three dimensions of the slice contain crucial feature information, particularly in the width and height dimensions, which are indispensable for the final image fusion. 

\noindent \textbf{Effectiveness of Dimensional Rolling Transformation Module.}
By comparing the results of groups (a), (h), and (i), it is evident that DRTM plays a crucial role in the final \mbox{fusion} outcome.  
The removal of DRTM leads to a performance decrease of nearly 2 dB in terms of PSNR, highlighting the critical role and effectiveness of our design.

\noindent \textbf{Effectiveness of Attention Cache.}
Based on the results in Table~\ref{table5}, although performance is slightly reduced due to quantization compression in the attention cache, the inference time improves significantly. 
Specifically, it decreases from 0.1475 seconds to \textbf{0.0441} seconds, the processing time has reduced by 70\%, enabling real-time inference.

\section{Discussion}
\noindent \textbf{Discussion of Chunk-Cache Pipeline.}
To investigate the necessity and effectiveness of the chunk-cache pipeline, we conduct it to the comparison method AHDRNet~\cite{c:8} and conduct experiments on our proposed 4K-DMEF dataset. 
Specifically, we divide the input images into blocks of size 64 $\times$ 64 in both width and height dimensions. We then cache the intermediate results of each chunk, making them readily available for inference when needed. 
We show the result in Table~\ref{chart}. 
\begin{table}[t]
	\centering
	
	\begin{tabularx}{0.45\textwidth}{c| c c c}
		\Xhline{1pt}
		\textbf{Model} & \textbf{PSNR}$\boldsymbol{\uparrow}$ & \textbf{SSIM}$\boldsymbol{\uparrow}$ & \textbf{LPIPS}$\boldsymbol{\downarrow}$ \\
		\Xhline{1pt}
		AHDRNet & 20.04 & 0.5012& 0.4906 \\
		AHDRNet with cache & \textbf{22.81}& \textbf{0.6583} & \textbf{0.3529} \\
		\Xhline{1pt}
	\end{tabularx}
	\caption{Comparison between the AHDRNet and AHDRNet with the cache.}
	\label{chart}
\end{table}

\noindent \textbf{Discussion of Compression Methods.} 
For compression methods, we consider the following two approaches: 
Quantization compression, and convolutional neural network compression.
Table~\ref{discussion2} demonstrates that with quantization compression, the model's average inference time to generate one image is 0.0441 seconds.
Details on convolutional neural network compression and quantized compression can be seen in the Supplementary Material.

\begin{table}[t]
	\centering
	\begin{tabularx}{0.42\textwidth}{c| c c c}
		\Xhline{1pt}
		\textbf{Metrics} & \textbf{QC} & \textbf{CNN-W} & \textbf{CNN-C} \\
		\Xhline{1pt}
		Inference Time& \textbf{0.0441s} & 0.0690s& 0.1978s \\
		\Xhline{1pt}
	\end{tabularx}
	\caption{Comparison of compression methods. 
		QC refers to quantization compression, CNN-W denotes using CNN to compress the width dimension, and CNN-C denotes using CNN to compress the channel dimension.
	}
	\label{discussion2}
\end{table}

\noindent \textbf{Discussion of the Model's Potential.}
In this paper, We explore an case involving one \textbf{extremely overexposed} and one \textbf{extremely underexposed} input, which further validates the effectiveness of our model.  
Details are in the \mbox{Supplementary Materials}.

\section{Conclusion}
In this paper, we propose an advanced ultra-high-definition dynamic multi-exposure image fusion method via infinite pixel learning. 
Our approach features a chunk-cache-quantization pipeline that efficiently captures dimensional local features, avoids redundant computation, and accelerates inference on resource-limited devices. 
We also introduce a dimensional rolling transformation operation to associate and fuse different views of the feature map.  
Additionally, we provide a new UHD benchmark to evaluate the effectiveness of our method. 
Quantitative and qualitative results show that our proposed algorithm, can reach real-time ($>$40fps) for a single UHD image and generate satisfactory visual results on real-world UHD images.

\section{Acknowledgments}

\bibliography{aaai25}

\begin{thebibliography}{31}
\providecommand{\natexlab}[1]{#1}

\bibitem[{Chen et~al.(2024)Chen, Chen, Wu, Zheng, Pan, and Fu}]{UHDderaining}
Chen, H.; Chen, X.; Wu, C.; Zheng, Z.; Pan, J.; and Fu, X. 2024.
\newblock Towards Ultra-High-Definition Image Deraining: A Benchmark and An
  Efficient Method.
\newblock arXiv:2405.17074.

\bibitem[{Deng et~al.(2021)Deng, Ren, Yan, Wang, Song, and Cao}]{U:17}
Deng, S.; Ren, W.; Yan, Y.; Wang, T.; Song, F.; and Cao, X. 2021.
\newblock Multi-Scale Separable Network for Ultra-High-Definition Video
  Deblurring.
\newblock In \emph{ICCV}.

\bibitem[{Ge et~al.(2024)Ge, Zhang, Liu, Zhang, Han, and Gao}]{L:2}
Ge, S.; Zhang, Y.; Liu, L.; Zhang, M.; Han, J.; and Gao, J. 2024.
\newblock Model Tells You What to Discard: Adaptive KV Cache Compression for
  LLMs.
\newblock In \emph{ICLR}.

\bibitem[{Han et~al.(2024)Han, Wang, Xiong, Chen, Ji, and Wang}]{L:1}
Han, C.; Wang, Q.; Xiong, W.; Chen, Y.; Ji, H.; and Wang, S. 2024.
\newblock {LM}-Infinite: Simple On-the-Fly Length Generalization for Large
  Language Models.
\newblock In \emph{ICLR}.

\bibitem[{He et~al.(2016)He, Zhang, Ren, and Sun}]{He_2016_CVPR}
He, K.; Zhang, X.; Ren, S.; and Sun, J. 2016.
\newblock Deep Residual Learning for Image Recognition.
\newblock In \emph{CVPR}.

\bibitem[{Kalantari and Ramamoorthi(2017)}]{Kalantari2017DeepHD}
Kalantari, N.~K.; and Ramamoorthi, R. 2017.
\newblock Deep high dynamic range imaging of dynamic scenes.
\newblock \emph{ACM TOG}, 36: 1 -- 12.

\bibitem[{Li et~al.(2023)Li, Guo, man zhou, Liang, Zhou, Feng, and Loy}]{U:18}
Li, C.; Guo, C.-L.; man zhou; Liang, Z.; Zhou, S.; Feng, R.; and Loy, C.~C.
  2023.
\newblock Embedding Fourier for Ultra-High-Definition Low-Light Image
  Enhancement.
\newblock In \emph{ICLR}.

\bibitem[{Li et~al.(2020)Li, Ma, Yong, and Zhang}]{c:4}
Li, H.; Ma, K.; Yong, H.; and Zhang, L. 2020.
\newblock Fast Multi-Scale Structural Patch Decomposition for Multi-Exposure
  Image Fusion.
\newblock \emph{IEEE TIP}, 29: 5805--5816.

\bibitem[{Li et~al.(2024)Li, Yu, Liang, He, Karampatziakis, Chen, and
  Zhao}]{L:5}
Li, Y.; Yu, Y.; Liang, C.; He, P.; Karampatziakis, N.; Chen, W.; and Zhao, T.
  2024.
\newblock LoftQ: LoRA-Fine-Tuning-Aware Quantization for Large Language Models.
\newblock In \emph{ICLR}.

\bibitem[{Liu et~al.(2024)Liu, Gong, Wei, Dong, Cai, and Zhuang}]{L:6}
Liu, J.; Gong, R.; Wei, X.; Dong, Z.; Cai, J.; and Zhuang, B. 2024.
\newblock QLLM: Accurate and Efficient Low-Bitwidth Quantization for Large
  Language Models.
\newblock In \emph{ICLR}.

\bibitem[{Liu et~al.(2023)Liu, Zhang, Sun, Liang, Zeng, and
  Zhang}]{Mobile-HDRcvpr23}
Liu, S.; Zhang, X.; Sun, L.; Liang, Z.; Zeng, H.; and Zhang, L. 2023.
\newblock Joint HDR Denoising and Fusion: A Real-World Mobile HDR Image
  Dataset.
\newblock In \emph{CVPR}.

\bibitem[{Ma and Wang(2015)}]{c:2}
Ma, K.; and Wang, Z. 2015.
\newblock Multi-Exposure Image Fusion: A Patch-Wise Approach.
\newblock In \emph{IEEE International Conference on Image Processing}.

\bibitem[{Prabhakar et~al.(2021)Prabhakar, Senthil, Agrawal, Babu, and
  Gorthi}]{c:10}
Prabhakar, K.~R.; Senthil, G.; Agrawal, S.; Babu, R.~V.; and Gorthi, R. K.
  S.~S. 2021.
\newblock Labeled from Unlabeled: Exploiting Unlabeled Data for Few-shot Deep
  HDR Deghosting.
\newblock In \emph{CVPR}.

\bibitem[{Ren et~al.(2019)Ren, Ying, Li, and Li}]{LECARM}
Ren, Y.; Ying, Z.; Li, T.~H.; and Li, G. 2019.
\newblock LECARM: Low-Light Image Enhancement Using the Camera Response Model.
\newblock \emph{IEEE Transactions on Circuits and Systems for Video
  Technology}, 29(4): 968--981.

\bibitem[{Tan et~al.(2023)Tan, Chen, Zhang, Wang, Kan, Zheng, Jin, and
  Chen}]{tip23}
Tan, X.; Chen, H.; Zhang, R.; Wang, Q.; Kan, Y.; Zheng, J.; Jin, Y.; and Chen,
  E. 2023.
\newblock Deep Multi-Exposure Image Fusion for Dynamic Scenes.
\newblock \emph{IEEE TIP}, 32: 5310--5325.

\bibitem[{Tel et~al.(2023)Tel, Wu, Zhang, Heyrman, Demonceaux, Timofte, and
  Ginhac}]{c:12}
Tel, S.; Wu, Z.; Zhang, Y.; Heyrman, B.; Demonceaux, C.; Timofte, R.; and
  Ginhac, D. 2023.
\newblock Alignment-Free HDR Deghosting with Semantics Consistent Transformer.
\newblock In \emph{ICCV}.

\bibitem[{Tolstikhin et~al.(2021)Tolstikhin, Houlsby, Kolesnikov, Beyer, Zhai,
  Unterthiner, Yung, Steiner, Keysers, Uszkoreit, Lucic, and Dosovitskiy}]{MPL}
Tolstikhin, I.; Houlsby, N.; Kolesnikov, A.; Beyer, L.; Zhai, X.; Unterthiner,
  T.; Yung, J.; Steiner, A.; Keysers, D.; Uszkoreit, J.; Lucic, M.; and
  Dosovitskiy, A. 2021.
\newblock MLP-Mixer: An all-MLP Architecture for Vision.
\newblock In \emph{NeurlPS}.

\bibitem[{Wang et~al.(2024)Wang, Pan, Wang, Fu, Liang, Wang, Wu, and
  Liu}]{U:20}
Wang, C.; Pan, J.; Wang, W.; Fu, G.; Liang, S.; Wang, M.; Wu, X.-M.; and Liu,
  J. 2024.
\newblock Correlation Matching Transformation Transformers for UHD Image
  Restoration.
\newblock In \emph{AAAI}.

\bibitem[{Wang et~al.(2023)Wang, Zhang, Shen, Luo, Stenger, and Lu}]{U:19}
Wang, T.; Zhang, K.; Shen, T.; Luo, W.; Stenger, B.; and Lu, T. 2023.
\newblock Ultra-High-Definition Low-Light Image Enhancement: A Benchmark and
  Transformer-Based Method.
\newblock In \emph{AAAI}.

\bibitem[{Wang et~al.(2004)Wang, Bovik, Sheikh, and Simoncelli}]{P-S-metrics}
Wang, Z.; Bovik, A.; Sheikh, H.; and Simoncelli, E. 2004.
\newblock Image Quality Assessment: from Error Visibility to Structural
  Similarity.
\newblock \emph{IEEE TIP}, 13(4): 600--612.

\bibitem[{Woo, Ryu, and Kim(2021)}]{tip21}
Woo, S.-M.; Ryu, J.-H.; and Kim, J.-O. 2021.
\newblock Ghost-Free Deep High-Dynamic-Range Imaging Using Focus Pixels for
  Complex Motion Scenes.
\newblock \emph{IEEE TIP}, 30: 5001--5016.

\bibitem[{Xiao et~al.(2023)Xiao, Fu, Liu, Wu, and Zha}]{De-Raining_Transformer}
Xiao, J.; Fu, X.; Liu, A.; Wu, F.; and Zha, Z.-J. 2023.
\newblock Image De-Raining Transformer.
\newblock \emph{IEEE TPAMI}, 45(11): 12978--12995.

\bibitem[{Xu, Ma, and Zhang(2020)}]{c:9}
Xu, H.; Ma, J.; and Zhang, X.-P. 2020.
\newblock MEF-GAN: Multi-Exposure Image Fusion via Generative Adversarial
  Networks.
\newblock \emph{IEEE TIP}, 29: 7203--7216.

\bibitem[{Yan et~al.(2019)Yan, Gong, Shi, van~den Hengel, Shen, Reid, and
  Zhang}]{c:8}
Yan, Q.; Gong, D.; Shi, Q.; van~den Hengel, A.; Shen, C.; Reid, I.; and Zhang,
  Y. 2019.
\newblock Attention-Guided Network for Ghost-Free High Dynamic Range Imaging.
\newblock In \emph{CVPR}.

\bibitem[{Yang et~al.(2024)Yang, Han, Gao, Hu, Zhang, and Zhao}]{L:4}
Yang, D.; Han, X.; Gao, Y.; Hu, Y.; Zhang, S.; and Zhao, H. 2024.
\newblock PyramidInfer: Pyramid KV Cache Compression for High-throughput LLM
  Inference.
\newblock In \emph{ACL ARR}.

\bibitem[{Ye et~al.(2024)Ye, Tao, Huang, and Li}]{L:3}
Ye, L.; Tao, Z.; Huang, Y.; and Li, Y. 2024.
\newblock ChunkAttention: Efficient Attention on {KV} Cache with Chunking
  Sharing and Batching.
\newblock In \emph{ICLR}.

\bibitem[{Zhang et~al.(2018)Zhang, Isola, Efros, Shechtman, and Wang}]{lpips}
Zhang, R.; Isola, P.; Efros, A.~A.; Shechtman, E.; and Wang, O. 2018.
\newblock The Unreasonable Effectiveness of Deep Features as a Perceptual
  Metric.
\newblock In \emph{CVPR}.

\bibitem[{Zhang and Cham(2010)}]{c:1}
Zhang, W.; and Cham, W.-K. 2010.
\newblock Gradient-Directed Composition of Multi-Exposure Images.
\newblock In \emph{CVPR}.

\bibitem[{Zhang et~al.(2024)Zhang, Zhang, Wu, Yan, and Zuo}]{BracketIRE}
Zhang, Z.; Zhang, S.; Wu, R.; Yan, Z.; and Zuo, W. 2024.
\newblock Exposure Bracketing is All You Need for Unifying Image Restoration
  and Enhancement Tasks.
\newblock arXiv:2401.00766.

\bibitem[{Zhen~Liu and Liu(2022)}]{c:11}
Zhen~Liu, B.~Z., Yinglong~Wang; and Liu, S. 2022.
\newblock Ghost-Free High Dynamic Range Imaging with Context-Aware Transformer.
\newblock In \emph{ECCV}.

\bibitem[{Zheng et~al.(2021)Zheng, Ren, Cao, Hu, Wang, Song, and Jia}]{U:16}
Zheng, Z.; Ren, W.; Cao, X.; Hu, X.; Wang, T.; Song, F.; and Jia, X. 2021.
\newblock Ultra-High-Definition Image Dehazing via Multi-Guided Bilateral
  Learning.
\newblock In \emph{CVPR}.

\end{thebibliography}

\end{document}